\documentclass[runningheads]{llncs}
\usepackage{graphicx}
\usepackage{booktabs}
\usepackage{amsmath}
\usepackage{subcaption}
\usepackage{soul}
\usepackage{tabularx}
\usepackage{color}

\definecolor{lightgreen}{rgb}{0.686,0.839,0.686}
\sethlcolor{lightgreen}

\begin{document}
\title{Evaluating AI for Law: Bridging the Gap with Open-Source Solutions }
\titlerunning{A Call for an Open-source Legal Language Model}
\author{Rohan Bhambhoria\inst{1,2}\orcidID{0000-0002-2597-670X} \and \\
Samuel Dahan\inst{1,2,3}\orcidID{0000-0002-1079-8998} \and \\
Jonathan Li\inst{2}\orcidID{0000-0002-7095-805X} \and \\
Xiaodan Zhu\inst{1,2}\orcidID{0000-0003-3856-3696}
}
\authorrunning{R. Bhambhoria et al.}
\institute{Queen's University, Kingston ON  K7L3N6, Canada \\
\email{\{r.bhambhoria, samuel.dahan, jxl, xiaodan.zhu\}@queensu.ca}\\
\and
Ingenuity Labs \\
 \and
Cornell University\\
}
\maketitle              % typeset the header of the contribution
\begin{abstract}
This study evaluates the performance of general-purpose AI, like ChatGPT, in legal question-answering tasks, highlighting significant risks to legal professionals and clients. It suggests leveraging foundational models enhanced by domain-specific knowledge to overcome these issues. The paper advocates for creating open-source legal AI systems to improve accuracy, transparency, and narrative diversity, addressing general AI's shortcomings in legal contexts.

\keywords{Law  \and Open-Source \and Large Language Models.}
\end{abstract}
\section{Introduction}

In recent times, the rise of Large Language Models (LLMs) has become prominent, especially with the unprecedented growth of ChatGPT, marking it as the fastest-expanding consumer application to date. LLMs have shown extensive utility in tasks related to productivity and in systems designed for low-stakes decision-making, such as composing emails. However, its application in high-stakes decision-making areas, including contract drafting or medical diagnostics, is met with cautious adaptation due to concerns about its large-scale deployment. AI models are \textit{notorious bullshitters} \cite{ref_article1}. 

We believe that Artificial Intelligence-powered legal advice, or "legal AI," can improve access to justice and contribute more broadly to the practice of law, increasing the percentage of adequately-represented litigants and driving down legal fees and the cost of legal research (Dahan and Liang 2020; Surden 2019). Research has shown that legal AI can successfully assist self-represented litigants with tasks such as determining the validity of a claim, online dispute resolution and court filing (Dahan et al. 2020; 2023). However, AI remains plagued with limitations, including the phenomenon of “hallucination”  and the potential for biased advice \cite{ref_merkin}; and there is growing evidence of reliance – even overreliance – on the use of generative AI tools for legal advice. Alarmingly, this is true among both self-represented litigants and legal practitioners themselves \cite{ref_merkin} (Merken 2023, Little 2024). 

The problem is further exacerbated by the proliferation of so-called AI applications specialised for law, many of which are actually general-purpose AIs papered over with a legal interface. These applications, misleadingly presented as specialized legal tools, mask the underlying limitations of the technology. It should be noted that some large legal database providers have introduced genuine generative AIs for law. However, these systems, which are closed, raise transparency and accessibility concerns; conflict with the movement towards open science; and hinder understanding of AI knowledge \cite{ref_rudin}. Furthermore, these technologies are accessible primarily by large firms, thus offering limited benefit to broader legal communities and doing nothing to improve access to justice. 

Finally, current regulations, such as the EU AI Act and Canada's AIDA (Artificial Intelligence and Data Act ),  do not yet provide solid benchmarks for addressing these issues. Policymakers need to debate the accessibility, quality, impact, and design of legal AI in order to produce industry standards for its creation and use. Anything less not only hinders access to justice, but risks hurting the very people who stand to gain the most from its development.

We propose two solutions to address the identified challenges: (i) revising benchmarks and protocols to  evaluating legal AI's performance capabilities and limitations in real-world settings, and (ii) based on performance findings, creating a domain-specific, open-source language model interface that allows for diverse feedback collection. 

Acknowledging the current AI regulations' inadequacies in managing generative AI's risks in legal contexts, we advocate for a sector-specific approach to define reliable legal AI systems. This includes developing benchmarks specifically tailored to legal challenges, focusing on legal misinformation, transparency, and diversity in narrative representation. Utilizing computational benchmarking, we aim to develop evaluation metrics centered on Bias Risk, Fact-Checking, Legal Reasoning Ability, and Narrative Construction Diversity for application in question-answering tasks. 

Drawing from performance insights, we explore legal AI solutions that may show significant improvement over general-purpose AI. Specifically, we highlight the potential of an open-source approach through www.OpenJustice.ai—a platform that encourages collaborative and crowdsourced efforts to design and test custom AI solutions for legal professionals and aid centers. This approach promotes a transparent and inclusive method of AI development, allowing diverse perspectives and expertise to contribute to more ethical and robust AI systems. 

An essential aspect of this domain-specific solution is the emphasis on data curation. The quality of data used to train these models is crucial for their reliability and effectiveness \cite{ref_article22}. Proper data curation entails not only gathering extensive and diverse datasets but also ensuring the data is representative, unbiased, and pertinent to the specific legal applications. 

\section{Background:  AI Is Not Yet Ready for Law } 
\label{implic}
% problem

Recent studies indicate a concerning trend in artificial intelligence: the as-yet-unexplained "drifting" phenomenon, characterized by significant fluctuations in AI’s capabilities (Chen, Zaharia, and Zou 2023). For example, in one study, an AI system’s accuracy rate in solving basic math problems dropped from 98\% to 2\% in the space of months. Certainly in the legal context, evidence has shown that despite AI’s capacity to perform some legal tasks—even passing the bar exam (Katz et al. 2023)—the technology has not yet fully matured. Generative AI is prone to “hallucinating” inaccurate responses with confidence, offering biased advice and erroneous citations (Chen et al. 2023). A further problem is that Large Language Models (LLMs) tend to reflect a mainstream worldview. When they are a primary source of AI training data, feedback loops are created wherein AI-generated texts are reincorporated into the web, creating “AI echo chambers” (Shur-Ofry 2023, 30).  

\noindent \textbf{1. Hallucinations, and confident regurgitation. }Recent studies have demonstrated that generative AI is capable of executing a variety of legal tasks and has even successfully passed the bar examination \cite{ref_article5}. However, the technology still has significant limitations. A well-documented issue with generative AI architectures is their tendency to produce \textit{hallucinated} responses. These responses are characterized by high confidence levels in presenting incorrect information, including the fabrication of facts, citations, and details. Generalized Large Language Models (LLMs) such as ChatGPT operate by predicting a sequence of words that logically follows an initial user-provided input. This process incorporates an element of creativity, wherein the model randomly selects elements of a sentence from a set of probable responses. This method, while innovative, also contributes to the challenge of ensuring accuracy and reliability in the generated content, particularly in contexts that demand high precision, like legal tasks.

In essence, AI systems primarily operate on statistical principles and thus possess limited comprehension capabilities, particularly in specialized domains such as law. The generative models currently in use demonstrate a notable deficiency in capturing the semantic subtleties inherent in legal terminology. This limitation is exemplified by the varying interpretations of the same legal term across different jurisdictions. For instance, the term \textit{layoff} is interpreted as \textit{suspension} in Canada, whereas it denotes \textit{termination} in the United States. A more pressing concern is the deceptive proficiency of generative AI systems in understanding legal concepts. Despite appearances, these systems often lack the capability for counterfactual legal reasoning or the classification of different modes of legal reasoning. This shortfall highlights the gap between the apparent and actual capabilities of AI in complex, context-dependent fields such as law \cite{ref_article6}.

This limitation may be inconsequential for tasks that are statistical in nature, such as the retrieval of legal precedents or the application of straightforward rules to facts. However, it presents a significant challenge for more complex legal reasoning tasks, which require a multidimensional approach and an in-depth understanding of legal issues. Beyond the problem of \textit{hallucinations} or the generation of inaccurate content, there are also concerns regarding biased analyses. This is a well-documented issue in the application of predictive AI in legal contexts, as referenced in \cite{ref_article7} and \cite{ref_article8}. For instance, ChatGPT, like many AI systems, has been shown to exhibit biases that are commonly observed in human reasoning. These include conjunction bias, probability weighting, overconfidence, framing effects, anticipated regret, reference dependence, and confirmation bias, as detailed in \cite{ref_article9}. Such biases can significantly affect the objectivity and reliability of AI-generated legal analyses, underscoring the need for cautious application and rigorous evaluation of these technologies in legal settings.

\noindent \textbf{2. Lack of diversity in generated responses. }The implications of employing Large Language Models (LLMs) in the realm of computational linguistics are not limited to issues of inaccuracies or the dissemination of misinformation in legal contexts. A significant concern arises even when generative AI systems like LLMs provide accurate information. These concerns stem from the inherent tendency of general-purpose LLMs to replicate dominant worldviews. If LLMs become a prevalent source of information, there is a potential for creating feedback loops. In such scenarios, the text generated by LLMs could re-enter the digital ecosystem, effectively becoming part of the training dataset for subsequent generations of text-generating models. This could lead to the development of \textit{AI echo chambers}, as articulated in \cite{ref_article4}. Such chambers could substantially limit the diversity of thought, posing a risk to the breadth of intellectual discourse and possibly affecting the development of AI systems in ways that could hinder cultural diversity, narrative plurality, and the dynamism of democratic discourse.

\noindent \textbf{3. Linear Reasoning. } In the development and application of Large Language Models (LLMs) for legal domains, several critical issues emerge. Unlike fields with clear, mathematical solutions, law encompasses a spectrum of acceptable answers and allows for considerable discretion. This characteristic renders the use of LLMs in legal contexts not merely as tools for information retrieval, but as systems that inherently shape the representation of legal information. The first concern with such LLMs is the presumption that legal issues are algorithmic, necessitating straightforward solutions. This assumption contradicts the reality of legal practice, where different lawyers and judges often propose divergent solutions to identical legal scenarios, as evidenced in sources \cite{ref_article10,ref_article7,ref_article11}. A common lawyer's response, \textit{it depends}, highlights this variability.

\noindent \textbf{4. Lack of temporal dimension in legal applications. }LLMs often operate on the assumption that the facts and contexts of future legal cases will mirror those of the past, disregarding the dynamic nature of social contexts that continuously introduce new facts and legal challenges. This approach risks stagnating the legal field, potentially leading to the ossification and de-norming of law.

\noindent \textbf{5. Static training databases. }A further complication arises from the opaqueness of generative AI systems like LLMs, which often cannot cite their information sources. This is particularly problematic in legal practice, where the substantiation of arguments with appropriate legal citations is crucial. The inability of LLMs to reliably cite sources, and their tendency to generate fictitious case law or legislation poses significant risks. While models like GPT-4 have shown capabilities to reference relevant statutes and legal authorities, they continue to exhibit limitations in generating accurate connections and in incorporating new legal developments, given their training on a static dataset representing a snapshot of the internet at a particular time.

Yet, despite these sizable concerns, there is little empirical data about AI performance in legal context. Most knowledge stems from GPT-4's varied performance on legal exams \cite{legal_exams,legal_exams2,legal_exams3} and discrete questions \cite{questions_1,questions_2}, alongside theoretical discussions on AI's ethical use \cite{ethical} and its impact on legal skills and firm competitiveness \cite{competitiveness}. Research reveals GPT-4's inconsistent exam results and its challenges in issue identification and rule application, noting its potential to assist lower-performing students without benefiting top achievers \cite{barexam_misleading}. Limited evidence exists outside exams, with some studies highlighting AI's potential in legal reasoning and others cautioning against overreliance due to AI's hallucinations and misinterpretations. 

In fact LegalBench \cite{legalbench} is a collaborative project designed to benchmark the legal reasoning capabilities of Large Language Models (LLMs) using the IRAC framework. Its goal is to assess how well current AI models can support and augment legal reasoning, particularly in administrative and transactional settings, without aiming to replace legal professionals. In contrast, our project diverges by focusing on applying AI to practical, real-world legal tasks through a domain-specific, open-source platform. We aim to directly evaluate AI's effectiveness in performing tasks that mirror the day-to-day work of legal professionals, moving beyond theoretical benchmarks to assess practical utility and integration into legal workflows.

 Contrastingly, recent work investigates AI's capabilities in contract review, a narrowly defined task where AI has been shown to excel, even outperforming lawyers in previous studies \cite{bettercallgpt,bibgeeks}. This focus contrasts with our project, which extends beyond document analysis to cover a broader range of practical legal tasks through a domain-specific, open-source platform, aiming to assess AI's utility in a wider legal context. 

 Finally, recent work by Choi et al,  presents meaningful insights into how AI may augment the performance of some lawyers, particularly the lower-performing ones. However, it also primarily focuses on tasks that are relatively easier for LLMs, such as drafting legal documents and answering hypothetical questions with provided materials. Our project aims to bridge these gaps by providing a more detailed and nuanced examination of AI's capabilities and deficiencies in a broader range of legal tasks, moving beyond the realms explored by the aforementioned studies. This endeavor not only highlights AI's current limitations in legal practice but also advocates for a targeted approach towards developing domain-specific, open-source legal AI solutions to address these challenges effectively.
 
Our preliminary study aims to advance the literature on AI  benchmarking focusing on evaluating GPT-4's assistance in practical lawyering \cite{barexam_misleading} tasks - especially Question-Answer, aiming to advance the understanding of AI's qualitative impacts in law.

 % There's a bias in models to assume false premises in queries are true, and even hallucination rates can be 69\%-88\% \cite{https://arxiv.org/abs/2401.01301} under various legal tasks. 
% \footnote{https://hai.stanford.edu/news/hallucinating-law-legal-mistakes-large-language-models-are-pervasive\?fbclid=IwAR3OINcdoxptTBKLI85p9iyls\_YFkDdQ\_NcM7FggUdy9M7Wm3B6tuB2zA2o}

\section{Datasets and Statistics}
\begin{figure}[t!]
    \begin{subfigure}[b]{1\textwidth}
    \centering
    \caption{LegalQA: Distributions of Sequence Lengths}
    \includegraphics[width=\linewidth]{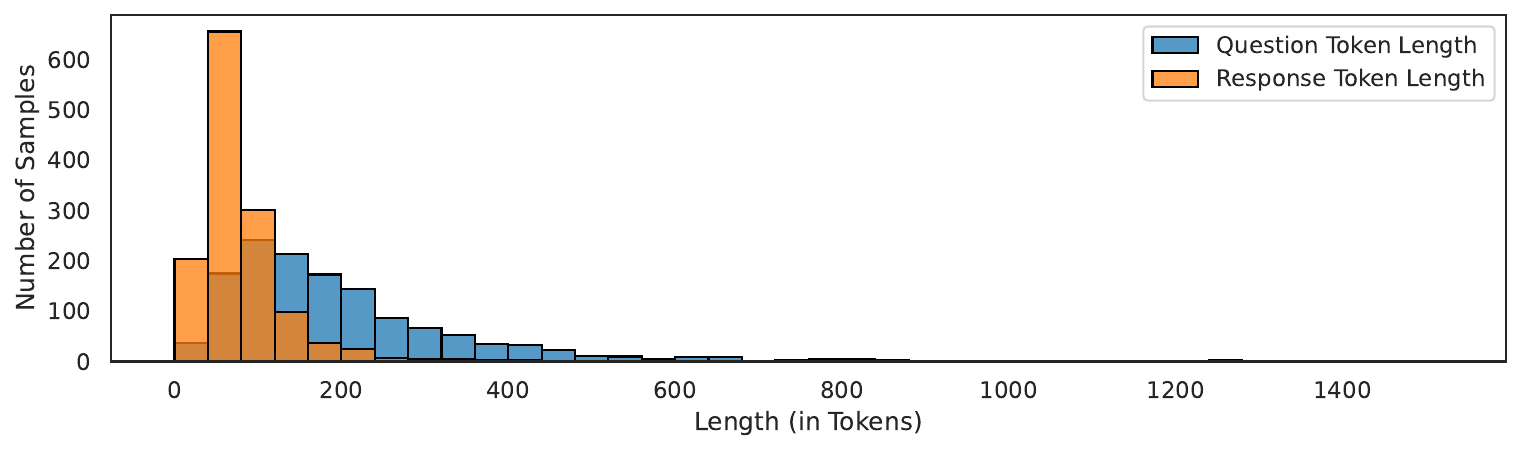}
    \label{fig:legalQAStats}
    \end{subfigure}
    \begin{subfigure}[b]{1\textwidth}
    \centering
    \caption{Law Stack Exchange: Distributions of Sequence Lengths}
    \includegraphics[width=\linewidth]{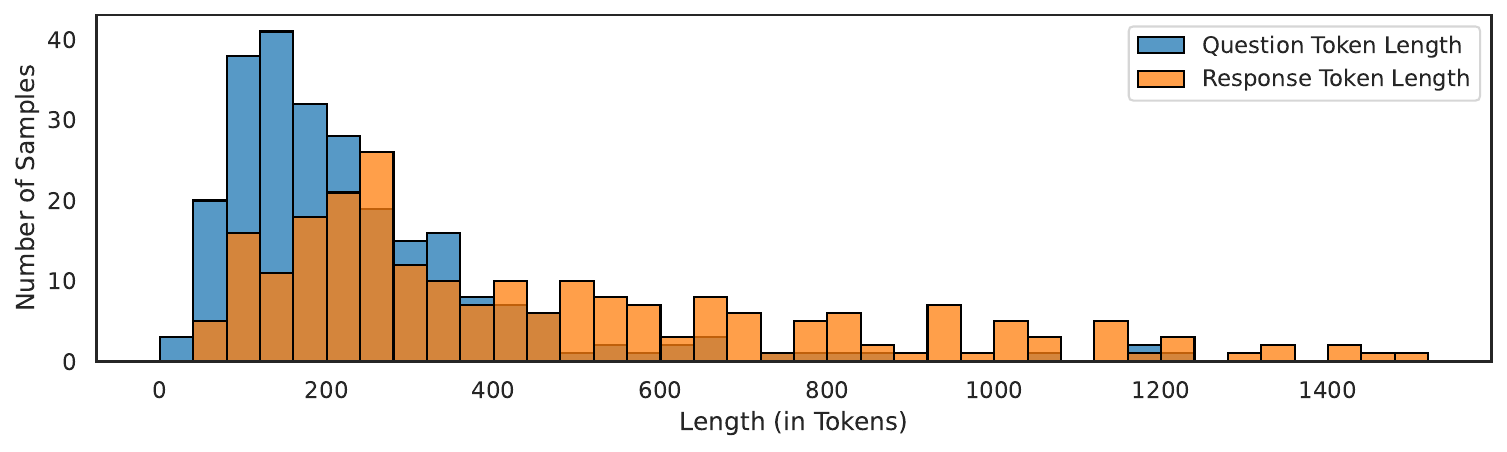}
    \label{fig:legalSEStats}
    \end{subfigure}
    \caption{Distribution of sequence lengths for LegalQA and Law Stack Exchange. We measure the length in tokens (with byte-pair encoding) and combine the train and test sets.} 
    \label{fig:sequenceLengths}
\end{figure}
In this work, we curate a legal QA benchmark to reflect real-world scenarios. Existing legal benchmarks, such as LegalBench \cite{ref_article21}, often lack real-world hypothetical and document analysis aspects present in real-world legal challenges. Specifically, most LegalBench tasks are classification problems, which fail to evaluate various dimensions desirable for a language model in aiding laypeople with legal tasks.

To this end, we curate LegalQA, a high-quality dataset of over 2000 questions asked by laypeople on real legal questions and answers vetted by legal experts. We ask law students to write expert answers to these questions (process described in more detail in Section~\ref{sec:annotation}). The questions are sourced from an online legal community\footnote{https://reddit.com/r/legaladvice}.

In conjunction with our new dataset, we also source additional legal questions from Law Stack Exchange and collect the 200 most popular questions on the site (to serve as a proxy for questions that people find interesting/challenging to solve). Then, we use the top-voted answer for each question as the expert label. 

These two datasets, from differing online platforms, address different domains. For example, the longer sequence lengths visible in Figure~\ref{fig:sequenceLengths} reflect differences in data collection methodologies; we encourage our law students to answer the question as concisely as possible (while addressing everything in the question), whereas top stack exchange answers often directly copy a long block quotation to cite their sources clearly, increasing length. Together, these two datasets provide an evaluation of a model's capabilities of addressing a layperson's legal concerns across various settings.

\subsection{Annotation Guidelines}
\label{sec:annotation}
We designed a data labelling procedure to build the most robust answers. Following principles described in Section~\ref{sec:openjustice}, we ask our human annotators to highlight relevant facts in the legal question, write a concise answer, and provide a link to a reliable source of further reading.

Since our law students are trained under Canadian law, we ask law students to consider questions on U.S. law from a Canadian perspective. If it is impossible to contextualize within a Canadian legal framework (e.g., if the question concerns specific gun laws), then we omit this data. Otherwise, if the question asks about the law that is answerable with Canadian law, then our annotators will treat it under a Canadian context. We consider these laws from a Canadian perspective to maintain annotations that are verifiable by our Canadian legal experts.

Regarding the format of answers, we ask annotators to limit answers to 200-300 characters, and to stay as close to the evidence as possible---this is to mirror a real-world legal environment where making concise and evidence-driven answers is important. Additionally, we ask annotators to avoid using legalese and explain in terms understandable by a general audience (the target audience). We include an example annotation in Table~\ref{tab:exampleData}.

\begin{table}[!t]
\centering
\small
\begin{tabularx}{\linewidth}{lX}
\toprule
Source & 
... my partner and I have been renting from a deadbeat property management company for the last year. There was a maintenance issue and a flooding issue 2 months after we moved in. We brought to their attention, they gave it a bandaid fix and have been ignoring any contact since (I have records of all of the attempts to contact them). We have called, texted, emailed, and my partner even went to their office to try and speak to someone and basically got blown off by the guy at the front desk who is one of the main agents for this company. Luckily, the issue hasn't caused damage to any of our personal property, but if it goes unfixed will inevitably damage the house itself and become a much bigger issue. Suffice to say, we're not interested in continuing to rent from them. The issue I'm seeking advice on is that \hl{we got an email asking if we're interested in renewing (for \$50 more a month for another 12 months)} and \hl{they are requesting we give notice by 3/1 if we intend to vacate}. \hl{Our lease we signed says we must provide notice by 3/31, not 3/1}. It's a difficult time all around to find a new place to rent that fits into our timeline and budget and I want as much time as we can so we don't have to overlap leasing and also to guarantee we can actually find something before our lease is up at the end of April. So my question is: should/can I send an email saying the lease we signed says we don't have to give notice until 3/31 whether or not we'll renew and that they can expect our answer then? Although they can request we give them more advanced notice, contractually, I don't think we're obligated to. Any advice would be greatly appreciated and thanks for taking the time to read this.
\\
\midrule
Provided answer & In Ontario, in a fixed-term tenancy, you must give at least a 60-day notice before the end of the lease. If not the tenancy continues. If the tenancy agreement does not say what happens after the end of the fixed-term, your lease will continue on a month-to-month basis automatically. A landlord cannot force you to agree to end a tenancy". \\
\midrule
Provided citation & \texttt{https://www.siskinds.com/pet-custody-laws-in-ontario} \\
\bottomrule
\end{tabularx}
\caption{Example question (``source'') and provided answers and citation. The highlighted green portion denotes the relevant facts indicated by the annotator.}
\label{tab:exampleData}
\end{table}

% hypothetical vs document analysis — how current tasks lack this
% how the dataset was curated
% community sources, reviewed by our lawyers

\label{sec:experiments}
\section{Experimental Setup}

To evaluate the landscape of current language models, we run experiments on state-of-the-art closed and open-sourced models. Specifically, we evaluate OpenAI's most recent GPT-4 model, GPT-4-Turbo-1106 (hereinafter called ``GPT-4'') since it is the state-of-the-art general-purpose language model at the time of writing. For the open-sourced model, we evaluated Mixtral-8x7B\footnote{https://mistral.ai/news/mixtral-of-experts/}, a state-of-the-art mixture of experts chat-aligned language model with 46.7B parameters.

We evaluate the factuality (as opposed to style) of the resulting generations. Since the real-world nature of this benchmark facilitates open-ended responses, it is difficult to evaluate model responses without human feedback. As a proxy for human evaluation, we evaluate the generated answers relative to the expert answer automatically with a language model, based on the opensource automatic evaluation repository, OpenAI Evals\footnote{https://github.com/openai/evals}. As OpenAI Evals were created for general-purpose evaluations, we would like to highlight the need and requirement for improvements in the legal domain. We use GPT-4 for this automatic evaluation, asking it to compare the model-generated answer to the expert answer, and indicating if the answer is a subset of the expert answer, a superset of the expert answer, contains the same details as the expert answer, disagrees with the expert answer, or answers the question in a incomparable way to the expert answer (see Table~\ref{tab:classes}). Then, we ask legal experts to review the model's evaluations to assess their reliability.

\begin{table}[!t]
\centering
\small
\begin{tabularx}{\linewidth}{lX}
\toprule
Category & Description \\
\midrule
A ($\subset$) & The submitted answer is a subset of the expert answer and is fully consistent with it. \\
B ($\supset$) & The submitted answer is a superset of the expert answer and is fully consistent with it. \\
C (=) & The submitted answer contains all the same details as the expert answer. \\
D ($\neq$) & There is a disagreement between the submitted answer and the expert answer. \\
E ($\overset{?}{\approx}$) & It is not possible to compare the answers directly for factuality because the submitted answer addresses the question differently from the expert answer. \\
\bottomrule
\end{tabularx}
\caption{Classes used to automatically evaluate the factuality of model generations.}
\label{tab:classes}
\end{table}

\begin{figure}[!t]
    \centering
    \begin{subfigure}{0.49\textwidth}
    \centering
    \caption{LegalQA}
    \includegraphics[width=\linewidth]{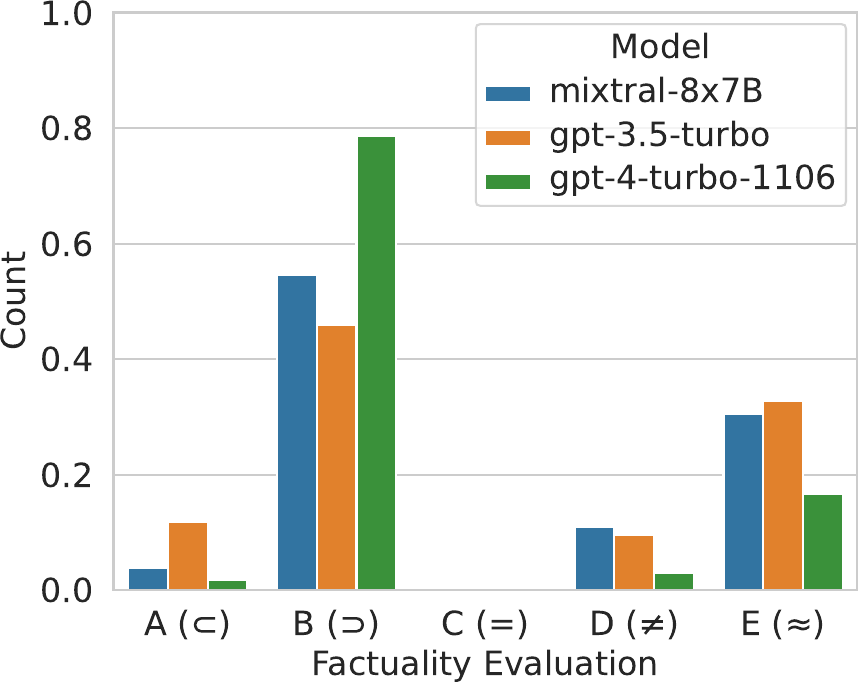}
    \label{fig:legalQAStats}
    \end{subfigure}
    \hfill
    \begin{subfigure}{0.49\textwidth}
    \centering
    \caption{Law Stack Exchange}
    \includegraphics[width=\linewidth]{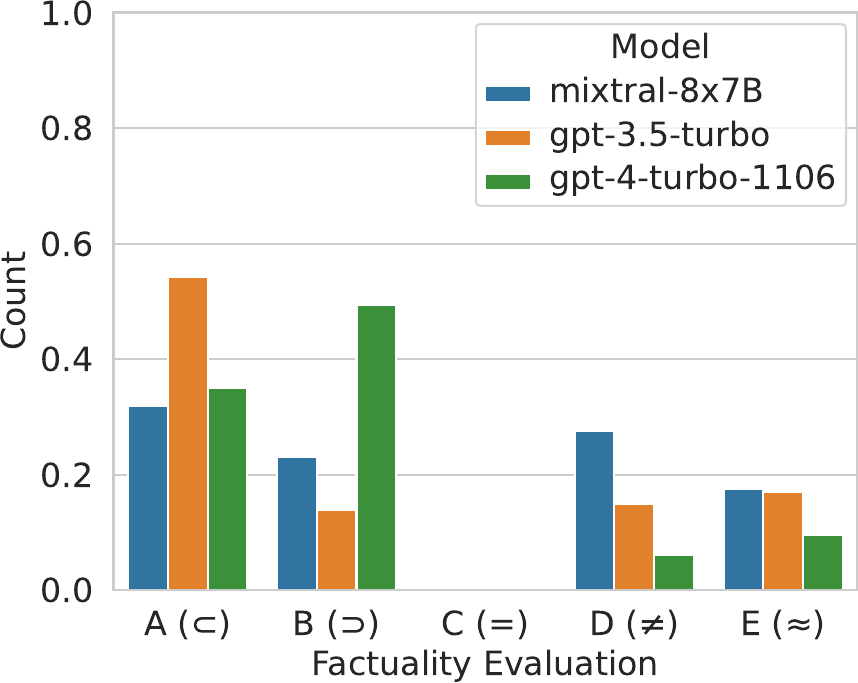}
    \label{fig:legalSEStats}
    \end{subfigure}
    
    \caption{Automatic evaluation results for (\subref{fig:legalQAStats}) LegalQA and (\subref{fig:legalSEStats}) Law Stack Exchange. Experimental setting described in Section~\ref{sec:experiments}}
    \label{fig:results}
\end{figure}

Our experiments reveal two insights. First, they evaluate various models in legal question answering. Second, they shed light on the capabilities of language models to be used for evaluation---something that is now done regularly in the general domain but underexplored in the legal domain.

\section{Results and Discussion}

In this section we conduct  experiments showcasing the ability of large language models in a practical legal question-answering task. Different from benchmarks {cite legal bench}, our task reflects real-world experimental settings.

Apparent in Figure~\ref{fig:results}, the state-of-the-art language model GPT-4 seems to perform relatively well on the LegalQA task, with under 5\% of examples containing factually incorrect responses. However, we observe that Mixtral 8x7B, a state-of-the-art open language model, falls significantly behind.

Additionally, we hypothesize that a large language model does not evaluate legal texts in the same way as a skilled human lawyer, so we ask law students to evaluate some predictions. The comments as a result of these qualitative observations are shown in Table~\ref{tab:feedback}. In general, we observe two phenomena that suggest that, despite the seemingly strong performance of GPT-4, there is still room for improvement:

\begin{itemize}
  \item \textbf{Lack of Citations.} When building a language model; the lack of credible citations given by models like GPT-4 stood out to our annotators---with an open model, it is easier to augment these models with tools, facilitating more trust in the form of citations.
  \item \textbf{Long Winded Answers.} Our annotators found answers written by humans to be more ``to the point'' and ``direct'', whereas GPT-4 often provides details that are not related to the legal question. For the examples evaluated, GPT-4 fails to capture the concise answers that legal experts are trained to write.
\end{itemize}

In the legal domain, these concerns---lack of citations and lack of concision---are especially important due the the high-stakes nature of legal decision-making. Given that the tasks tested---LegalQA and Law Stack Exchange---were relatively simple benchmarks (but still ahead of existing benchmarks), GPT-4's performance raises flags, indicated by comments in Table \ref{tab:feedback} on using these models for even more complex tasks, such as document analysis. 

Our results challenge the conclusions of  previous studies insofar as it shows that when LLMs, like GPT-4, are subjected to more complex legal reasoning tasks beyond document processing or answering questions with provided material, their performance diminishes. Unlike these studies that highlighted AI's efficiency in relatively straightforward tasks, our experiments reveal limitations in AI's ability to match the nuanced understanding and precise reasoning of human lawyers. Particularly, our findings underscore the issues of inadequate citations and verbosity in AI-generated responses, highlighting the necessity for domain-specific enhancements to meet the rigorous demands of legal practice.
\begin{table}[!t]
\centering
\small
\begin{tabularx}{\linewidth}{X}
\toprule
Comments \\
\midrule
I prefer the human answer in this particular circumstance. I thought the human answer was more direct an answering the question. It seemed as GPT-4 was providing plenty of details that were not necessarily relevant to overall answer to the question. The human answer also had at least some citations vs GPT-4. Neither opinion really provides much of an analysis of the circumstances.\\
\midrule
I prefer the human version. I think the GPT-4 answer is better written, but the human response provides an important citation and answers the question very directly. GPT-4 does not provide the same important citation. \\
\midrule
In this case, the specific example given by the human is perfect. 
It answers the question directly and shows the principles in play by example. GPT-4, while referring more to the example, is quite vague and long-winded in summarising that robbing a bank is stealing. \\
\midrule
Human response is far more to the point and interesting.
The GPT-4 answer starts by explaining the difference between crimes and civil wrongs without explaining why that's relevant from the start.
It could have gone directly into addressing the employer/employer theft concept, while integrating those ideas in. That is more so what the human response offered. The human also gave interesting context and history by mentioning how in certain jurisdictions debt merited imprisonment, or Australia considers wage theft a crime. It gives far more food for thought and is more of what the user was looking for. GPT-4 refers to labour law, but as it doesn't dive in with specifics, its quite vague and bland on the historical aspect.  \\
\midrule
Both specify there is no policy to prioritise the President of the US in organ transplantation. The human answer provides links to pages informing on the frameworks and mechanisms of organ transplantation. The GPT-4 answer considers the hypothetical case a little but remains vague. \\
\midrule
The human answer is superior as it directly references the relevant laws and rules, providing a more precise and targeted response. In contrast, GPT's answer appears somewhat vague, lacking explicit mention of the applicable rules, which could compromise the clarity and specificity of the information provided. \\
\bottomrule
\end{tabularx}
\caption{Feedback from law students on the automatic evaluations given by GPT-4.}
\label{tab:feedback}
\end{table}
\section{A Framework for Legal AI}

\subsection{The State of Legal AI}

Our results suggest that  legal industry needs domain-specific solutions.  Such systems should be able to handle complex legal reasoning and nuance: after all, in the legal sphere, questions often do not have a single “right answer”, but rather a range of acceptable answers. However, the past few years have seen an explosion of so-called generative AIs for law, many of which are actually general-purpose AIs hidden beneath legal-interface veneers. These systems obscure their underlying limitations, despite being presented as specialized tools. 

Unfortunately, consensus on the essential criteria for developing legal AI has not yet been reached. There are three primary strategies being considered. One possibility (Option 1) is to develop a purpose-built, law-specific generative AI model from the ground up. To our knowledge, no such initiative has yet been undertaken in law. It has been undertaken in finance with BloombergGPT (Wu et al. 2023); but data on its short- and long-term performance is limited, and recent evidence shows that finance-customized BloombergGPT does not outperform generic GPT4 (Li et al. 2023). To our knowledge, no such initiative has been undertaken in law, likely due to the cost. Furthermore, there is still limited data on short- and long-term performance, and recent evidence shows that BloombergGPT does not perform better than GPT4, a generic model (Li et al. 2023). A more economical choice (Option 2) involves fine-tuning an existing LLM, such as GPT or LLAMA, for legal applications. Early findings indicate that this could be an adequate short-term solution for some legal tasks (Guha et al. 2023). Finally, a third option (Option 3) entails training a Small Language Model (SLM) from scratch. Research shows that small-language models, such as Orca 2, can exhibit strong reasoning abilities and outperform larger models by learning from detailed explanation traces (Hughes 2023).

Irrespective of the selected approach, the essential requirement across all solutions  requires expertise in both law and computer science. In fact, computational law expertise is vital for constructing, evaluating, and refining legal AI systems. Engaging with professionals possessing this dual expertise is crucial, as they can provide nuanced and informed feedback essential for the continuous improvement of these systems.

However, building an effective legal AI tool extends beyond just expert involvement. Traditional approaches, such as creating closed-system AI platforms accessible only to law firms or those operating on restricted databases, are insufficient. Such approaches, exemplified by platforms like Harvey\footnote{https://www.harvey.ai/} and LexisNexis's Lexis Plus\footnote{https://go.lexisnexis.ca/lexis-plus-ppc-ai}, fall short in addressing the complexities highlighted in Section \ref{implic}. The limitations of these closed systems, primarily their lack of broad user engagement, can lead to stagnation, failing to adapt to the evolving legal landscape.

\subsection{OpenJustice: A Recipe for Building a Crowdsourced Legal Language Model}
\label{sec:openjustice}

% solution
To overcome these limitations, we propose the development of an open-access legal AI model. Launched in March 2023 by the Conflict Analytics Lab, OpenJustice entered development in January 2021 and operates as a natural-language processing interface, including question-answering, document analysis, and citation retrieval. The system has three layers of data. The first is a core open-source component trained on curated legal data, encompassing thousands of annotated question-answer pairs and case law from the US, Canada, France, and the EU. The second layer relies on crowdsourced human feedback, with OpenJustice handling approximately  a few thousands requests per week. Finally, the third layer uses proprietary data from partners to create custom models.  

Note that we are experimenting with LLM-fine tuning (Option 2) and training Small Language Models (Option 3). These models are accessible to a wider range of partners, such as law schools, aligning with goals like improving access to justice and legal education. More importantly, this open-access approach can significantly enhance the quality and effectiveness of the AI system. It addresses key issues such as the prevention of AI hallucinations, ensuring factual accuracy, promoting diversity of perspectives, and countering the static nature of traditional training databases. This open version, however, should be restricted to \textit{sophisticated users}{\textemdash}defined here as individuals with a legal background. This restriction is proposed to maintain the quality of the feedback loop. Sophisticated users, with their specialized knowledge and experience, are more likely to provide constructive and relevant feedback, which is crucial for the iterative improvement of the system. By incorporating inputs from a diverse yet expert user base, the AI system can be continuously refined and updated, ensuring it remains relevant and effective in a rapidly evolving legal environment.

\begin{figure}[!t]
    \centering
    \includegraphics[scale=0.25]{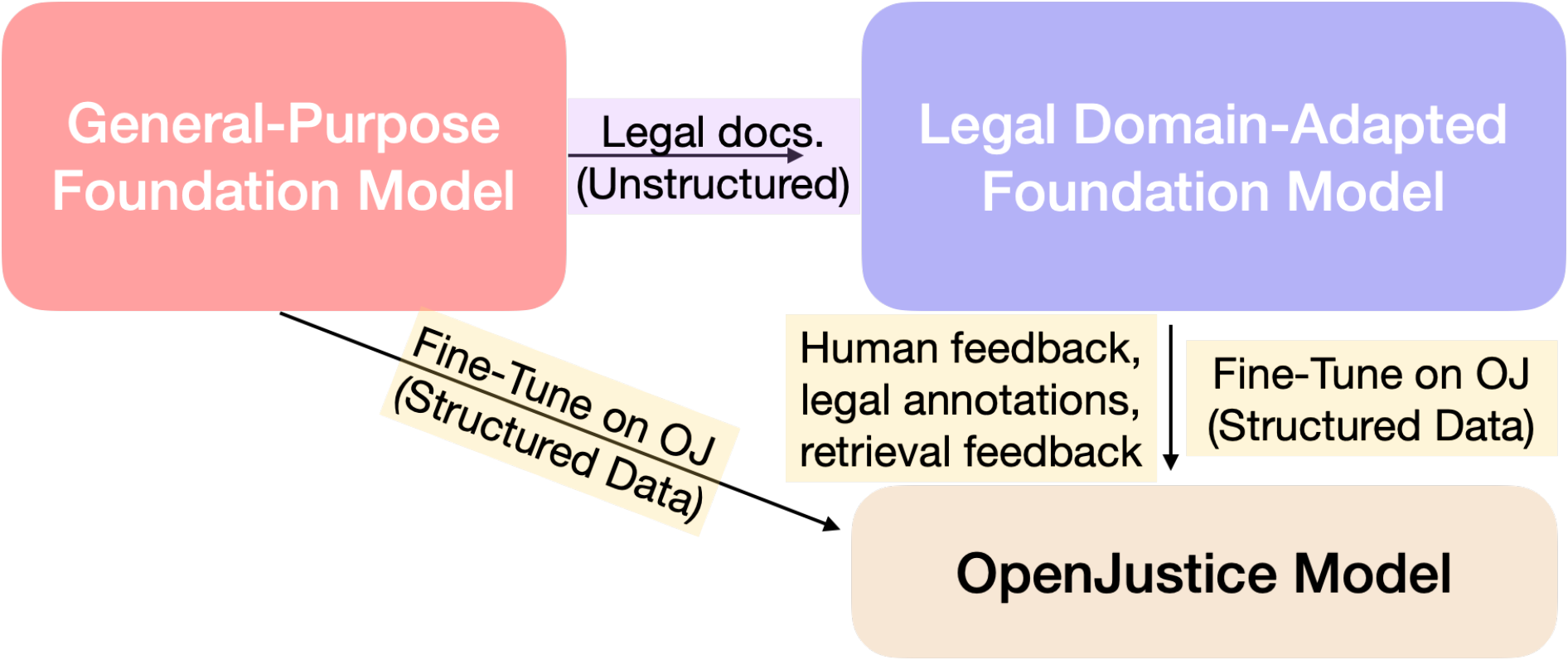}
    \caption{Legal Community Feedback utilized for OpenJustice}
    \label{fig:1}
\end{figure}

Note that feedback data provided by professionals will take many forms\textemdash(i) user behaviour, (ii) handwritten expert feedback, (iii) conversation steering, (iv) response ratings, (v) external feedback. All forms of data contribution by user-contributors, depicted in Figure \ref{fig:1} through a platform called \textit{OpenJustice\footnote{https://openjustice.ai/}}, are a valuable resource, which can be useful for supervision in our framework. Additionally, databases in the unstructured format may potentially be valuable by incorporating in our framework through pretraining or domain-adaptation. However, the value of the aforementioned remains unexplored in the legal domain, and has been shown to not be useful in other domains such as finance\cite{ref_article18}.

In the rapidly evolving field of artificial intelligence, particularly in the context of developing AI models for legal applications, a suite of advanced methodologies and research breakthroughs offer promising avenues. Direct Preference Optimization (DPO) \cite{ref_article19}, is particularly pertinent for legal AI due to its capability to align model outputs with complex, often subjective human preferences, a common characteristic in legal reasoning. This method's ability to refine AI outputs based on nuanced user feedback makes it exceptionally suited for the legal domain where interpretations and judgments are not always clear-cut. World models, as outlined in \cite{ref_article12}, offer another compelling tool for legal AI. These models enable the AI to create internal simulations of possible scenarios, an approach that mirrors the predictive and hypothetical reasoning often employed in legal analysis. By simulating various legal outcomes and scenarios, a world model-equipped legal AI can provide more comprehensive and informed suggestions. Advances like Flash Attention 2 \cite{ref_article20} contribute significantly by enabling the processing of large volumes of legal texts efficiently. This efficiency is crucial in legal contexts, where the ability to rapidly parse through extensive legal documents and case laws is a valuable asset. Furthermore, techniques like rejection sampling and reward modeling are instrumental in refining the quality of AI-generated legal content. Rejection sampling can be used to filter out less relevant or lower-quality content, ensuring that the AI's outputs are pertinent and of high quality. Reward modeling aligns the AI's objectives with desired outcomes, an essential feature when the AI is required to navigate the complex landscape of legal ethics and standards. Lastly, supervised fine-tuning and alignment research are integral to the development of a legal AI model. Supervised fine-tuning allows the model to learn from specific, high-quality legal datasets, ensuring that its outputs are relevant and accurate. Alignment research, which focuses on ensuring that AI systems' goals are aligned with human values and intentions, is particularly critical in the legal domain, where the stakes of misalignment can be particularly high.

 The integration of the aforementioned methodologies and research developments presents a comprehensive approach for training legal AI models. This integration not only enhances the capability of AI in understanding and processing legal information but also ensures that the AI's functioning is aligned with the nuanced, complex, and ethically bound nature of legal reasoning. Also,  this proposed open-access, yet expert-restricted, legal AI model represents a paradigm shift in the development of legal technology. It not only promises enhanced performance in terms of accuracy and relevance but also embodies a progressive approach towards democratizing legal knowledge and fostering an inclusive legal tech ecosystem.

\section{Conclusion}

Artificial intelligence has been bringing a profound impact on legal applications. In this work, we bring a call for open-source legal language models. This is done as a means to address limitations of general-purpose language models which may have underlying limitations, affecting their usage for high-stakes decision making. We provide a recipe for creating a legal language model, called OpenJustice. We further introduce a real-world high-quality dataset manually annotated by legal experts, called LegalQA, and run experiments outlining performance of current state-of-the-art LLMs. We place further emphasis on highlighting the limitations in the automatic evaluation process when used on legal datasets. 

\section{Acknowledgements}

We would like to thank students of the Conflict Analytics Lab at Queen's University Faculty of Law for being a part of initial efforts put into this initiative, and David Liang for coordinating efforts.

%
% the environments 'definition', 'lemma', 'proposition', 'corollary',
% 'remark', and 'example' are defined in the LLNCS documentclass as well.
%


\end{document}